\documentclass[english]{lni}

\usepackage{graphicx}
\usepackage{fancyhdr}
\usepackage{changepage} 
\usepackage{listings} 

\usepackage[figurename=Fig., tablename=Tab., small]{caption}

\fancypagestyle{titlepage}{
\fancyhead[RO]{\small F.  Boutros,  N. Damer,  M. Fang,  M. Gomez-Barrero,  K. Raja,  \linebreak C. Rathgeb,   A.  Sequeira,  and M. Todisco (Eds.): BIOSIG 2024, \linebreak Lecture Notes in Informatics (LNI), Gesellschaft f\"ur Informatik, Bonn 2024} 
\fancyfoot{}}

\renewcommand{\headrulewidth}{0.4pt} 
\author{Wassim Kabbani \footnote{IIK, Info. Sec. and Comm. Technology, Gjovik, Norway, wassim.h.kabbani@ntnu.no} \, Kiran Raja \footnote{Department of Computer Science, Gjovik, Norway, kiran.raja@ntnu.no} \, Raghavendra Ramachandra \footnote{IIK, Info. Sec. and Comm. Technology, Gjovik, Norway, raghavendra.ramachandra@ntnu.no} \, Christoph Busch \footnote{IIK, Info. Sec. and Comm. Technology, Gjovik, Norway, christoph.busch@ntnu.no}}

\title{Demographic Variability in Face Image Quality Measures}
\begin{document}

\maketitle

\renewcommand{\refname}{References}
\setcounter{footnote}{2} 
\thispagestyle{titlepage}
\pagestyle{fancy}
\fancyhead{} 
\fancyhead[RO]{\small Demographic Variability in Face Image Quality Measures \hspace{25pt}  \hspace{0.05cm}}
\fancyhead[LE]{\hspace{0.05cm}\small  \hspace{25pt} Anonymous for Review}
\fancyhead[LE]{\hspace{0.05cm}\small  \hspace{25pt} W. Kabbani, K. Raja, R. Ramachandra, C. Busch}
\fancyfoot{} 
\renewcommand{\headrulewidth}{0.4pt} 

\begin{abstract}

Face image quality assessment (FIQA) algorithms are being integrated into online identity management applications. These applications allow users to upload a face image as part of their document issuance process, where the image is then run through a quality assessment process to make sure it meets the quality and compliance requirements. Concerns about demographic bias have been raised about biometric systems, given the societal implications this may cause. It is therefore important that demographic variability in FIQA algorithms is assessed such that mitigation measures can be created. In this work, we study the demographic variability of all face image quality measures included in the ISO/IEC 29794-5 international standard across three demographic variables: age, gender, and skin tone. The results are rather promising and show no clear bias toward any specific demographic group for most measures. Only two quality measures are found to have considerable variations in their outcomes for different groups on the skin tone variable.

\end{abstract}

\section{Introduction}
\label{sec:intro}

Facial biometric systems have become an integral part of our daily lives. We interact with them when we unlock our phones, cross an automatic border control (ABC) gate, gain access to a physical or virtual space, or apply for an identity document. These systems provide a seamless and convenient experience and have been shown to achieve high performance on several benchmark tests \cite{DARGAN2020Survey}. However, concerns about bias and demographic fairness have recently accompanied the deployment of these systems \cite{Drozdowski2020DemographicBias}, and algorithmic bias has been considered one of the important open challenges in biometrics \cite{ross2019some}.

Studies have shown that facial image datasets as well as facial biometric algorithms such as face detection, face recognition, and gender classification are biased toward certain demographic groups \cite{buolamwini2018gender, Terhorst2020Comparison, hazirbas2021casual}. Thus, questions of how we assess and mitigate any demographic biases in any proposed AI system, and particularly in this context, facial biometric systems, are very often raised \cite{Katina2022BiometricsAIBias}. To help address these concerns and to standardize how we measure and talk about demographic fairness, an international standard ISO/IEC 19795-10 on quantifying biometric system performance variation across demographic groups is currently under development \cite{ISO-IEC-DIS-19795-10-230928}.

The international standard ISO/IEC 2382-37 on biometrics vocabulary \cite{ISO-IEC-2382-37-220330} defines the term \textit{demographic differential} as the \textit{extent of difference in outcome of a biometric system across socially recognized sectors of the population}. Hence, we use the term \textit{demographic differential} instead of \textit{demographic bias} when referring to any variations in the outcome of the algorithms between different demographic groups. 

In this work, we study demographic differentials in the area of face image quality assessment (FIQA), particularly assessing the quality measures included in the recent version of the international standard ISO/IEC 29794-5 on Biometric Sample Quality—Part 5: Face Image Data \cite{ISO-IEC-29794-5-DIS-FaceQuality-240129}. In its most recent version (2024-07-02), the Face Analysis Technology Evaluation (FATE) report on Specific Image Defect Detection (SIDD) from NIST, which summarizes the evaluation results of face image quality algorithms submitted to its platform, includes a section on demographic differentials by "subject region of birth" and for only five quality measures \cite{Yang_Grother_Ngan_Hanaoka_Hom_2024}. Instead, we focus on demographic differentials across three demographic variables: age, gender, and skin tone, which is a recently proposed alternative to traditional race categories when exploring performance differentials \cite{Howard-SkinPhenotypes-TBIOM-2021}.

Face image quality assessment (FIQA) refers to the process of assessing the utility of a face image to face recognition systems \cite{Schlett-FIQA-LiteratureSurvey-CSUR-2021} as well as assessing the image against a set of regulatory specifications \cite{ICAO-PortraitQuality-TR-2018, ISO-IEC-39794-5-G3-FaceImage-191015}. The ISO/IEC 29794-5 standard defines a set of quality \textit{measures} that apply to a face image. It categorizes these \textit{measures} into two main groups: (1) \textit{unified quality score} which gives an overall assessment that is not necessarily correlated with a particular aspect of the image. (2) \textit{quality component} which assesses a specific aspect of the face image such as illumination uniformity, dynamic range, luminance mean, and background uniformity \cite{ISO-IEC-29794-5-DIS-FaceQuality-240129}.

To study demographic differentials across the skin tone variable, a method for categorizing human skin tones is needed. The commonly used method in the literature is the Fitzpatrick Skin Types (FST) Scale. This scale describes a person's skin type in terms of its response to ultraviolet radiation (UVR) exposure and groups skin tones into six categories \cite{ward2018clinical}. However, while it may be useful for dermatological use cases, it has been shown to have a tendency toward lighter skin tones because they have more UV sensitivity \cite{Howard-SkinPhenotypes-TBIOM-2021, thong2023skincolor}. In their study, Howard et al. report that FST is \textit{poorly predictive of skin tone and should not be used as such in evaluations of computer vision applications} \cite{Howard-SkinPhenotypes-TBIOM-2021}. To address issues in FST and to offer a more inclusive scale that can be used to assess and mitigate bias in computer vision systems, Google has recently released the Monk Skin Tone Scale (MST) \cite{Monk_2019}. Introduced by Dr. Ellis Monk and made open source by Google, the MST introduces a more inclusive 10-tone scale explicitly designed to represent a broader range of communities. A study conducted by Google to understand how well participants across diverse communities felt their own skin tone was represented within the scale reported that participants found the MST Scale to be more inclusive than the Fitzpatrick Scale and better at representing their skin tone \cite{Monk_2019}. Together with the scale, Google also released the Monk Skin Tone Examples (MST-E) dataset, meant as a reference dataset to train human annotators on producing consistent skin tone annotations according to the 10-point MST scale \cite{schumann2023consensus}.

The approach we follow in this work is to study the score distributions of all the FIQA measures included in the ISO/IEC 29794-5 standard on four different datasets and across the three chosen demographic variables, namely: skin tone, age, and gender. To the best of our knowledge, this is the first study on demographic differentials in FIQA that considers three demographic variables and all the quality measures included in the newest version of the ISO/IEC 29794-5 standard \footnote{Due to the limited space, we include only the plots for the most important measures and interesting results for each demographic variable. But results for all quality measures on all three demographic variables, along with the produced skin tone ground truth labels, will be made publicly available.}. Besides, this is the first work in FIQA, and facial biometrics in general, to use the newly released and more inclusive Monk Skin Tone Scale (MST) to study and report skin tone bias.

\section{Experimental Setup}
\label{sec:setup}

\subsection{Datasets}
\label{subsec:datasets}

We use four datasets in total for the evaluation: FRLL \cite{DeBruine2021}, FRGCv2 \cite{Phillips-OverviewFaceRecognitionGrandChallengeFRGC-CVPR-2005}, LFW \cite{Huang2007LFWTech}, and MST-E \cite{schumann2023consensus}. All four datasets include images of subjects with different skin tones, genders, ages, and races. The FRLL dataset features images taken in a controlled studio environment, while LFW has images taken in the wild. FRGC and MST-E have images taken indoors, outdoors, and with different lighting, poses, and expression conditions. The MST-E Dataset is meant to be a reference dataset for the Monk Skin Tone Scale (MST) such that human observers can use it to train on how to label subjects on this scale, thus it includes ground truth labels about skin tone for each of the subjects \cite{Monk_2019, schumann2023consensus}. The other datasets do not have ground-truth skin tone labels. The FRLL dataset has ground-truth labels for age, gender, and race. LFW has manually verified gender labels \cite{afifi2019afif4}.

To overcome the lack of ground truth age, gender, and race labels for some datasets, we use Face Attribute Classification (FAC) \cite{park2022fair} to extract these labels when they are missing. The predicted labels are averaged on all images of the same subject to obtain more accurate results. However, there is no reliable automated method for predicting the real skin tone labels, so to make sure we obtain credible results for the skin tone analysis, we manually label 902 subjects from two datasets according to the MST's guidance. As per the Monk Scale Tone guidance, we use the MST-E dataset as a reference, and we label all subjects in the FRLL dataset and 800 subjects in the LFW dataset. The subjects in the LFW dataset are selected based on those that have the largest number of images to make sure that we investigate as many images of the same subject as possible before giving them an MST scale value.

\begin{table*}[h]
  \centering
  \begin{tabular}{|l|c|c|c|c|}
    \hline
    Dataset & \#Images & \#Subjects & \#Skin Tone Labelled Subjects \\
    \hline
    MST-E & 887 & 19 & 19 \\
    \hline
    LFW & 12684 & 5556 & 800 \\
    \hline
    FRLL & 597 & 102 & 102 \\
    \hline
    FRGC & 18154 & 227 & - \\
    \hline
  \end{tabular}
  \caption{Overview of the evaluation datasets. The numbers are for the actual number of subjects and images used in the evaluation after discarding images where no face is detected.}
  \label{tab:example}
\end{table*}

\subsection{FIQA Algorithms}
\label{subsec:measures}

To evaluate the FIQA measures defined in ISO/IEC DIS 29794-5, we use the reference implementation in the Open Source Face Image Quality (OFIQ) framework \footnote{https://github.com/BSI-OFIQ/OFIQ-Project}. OFIQ provides implemented algorithms for all quality measures.

\section{Experiments and Results}
\label{sec:experiments}

\subsection{Skin Tone}

\begin{figure*}[h]
    \centering
    \includegraphics[width=\linewidth]{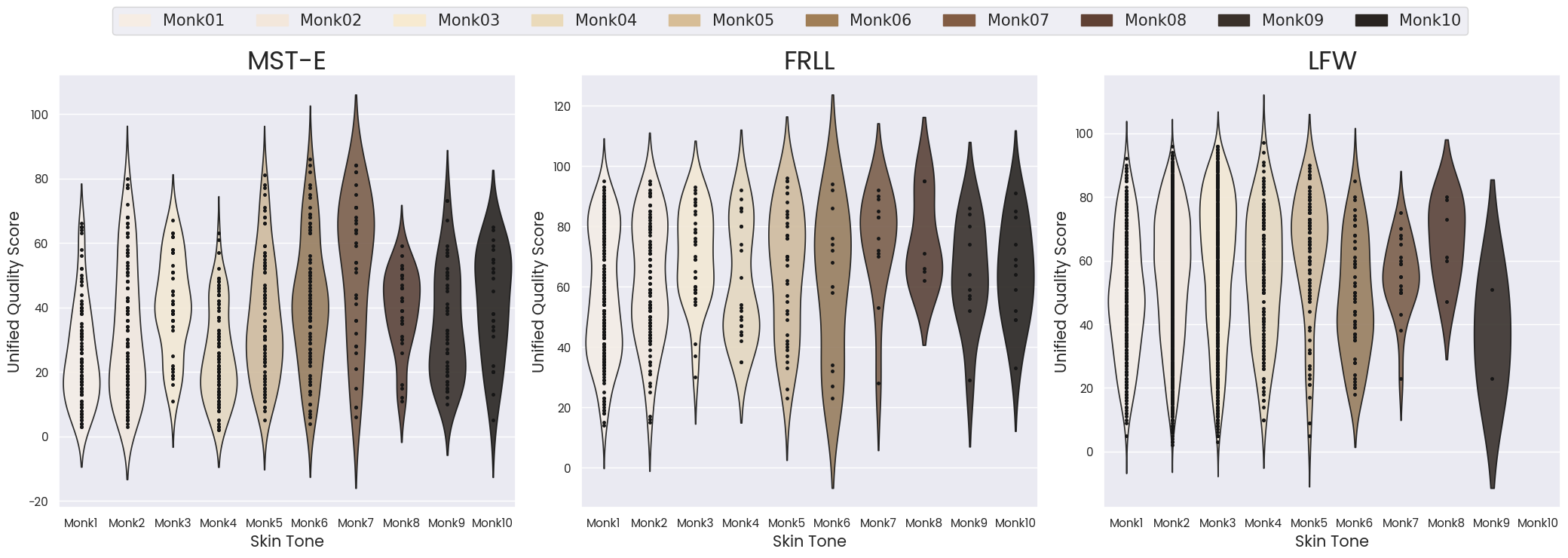}
    \caption{Unified quality score distributions across the MST 10 skin tone scale.}
    \label{fig:skintone-uqa}
\end{figure*}

\begin{figure*}[h]
    \centering
    \includegraphics[width=\linewidth]{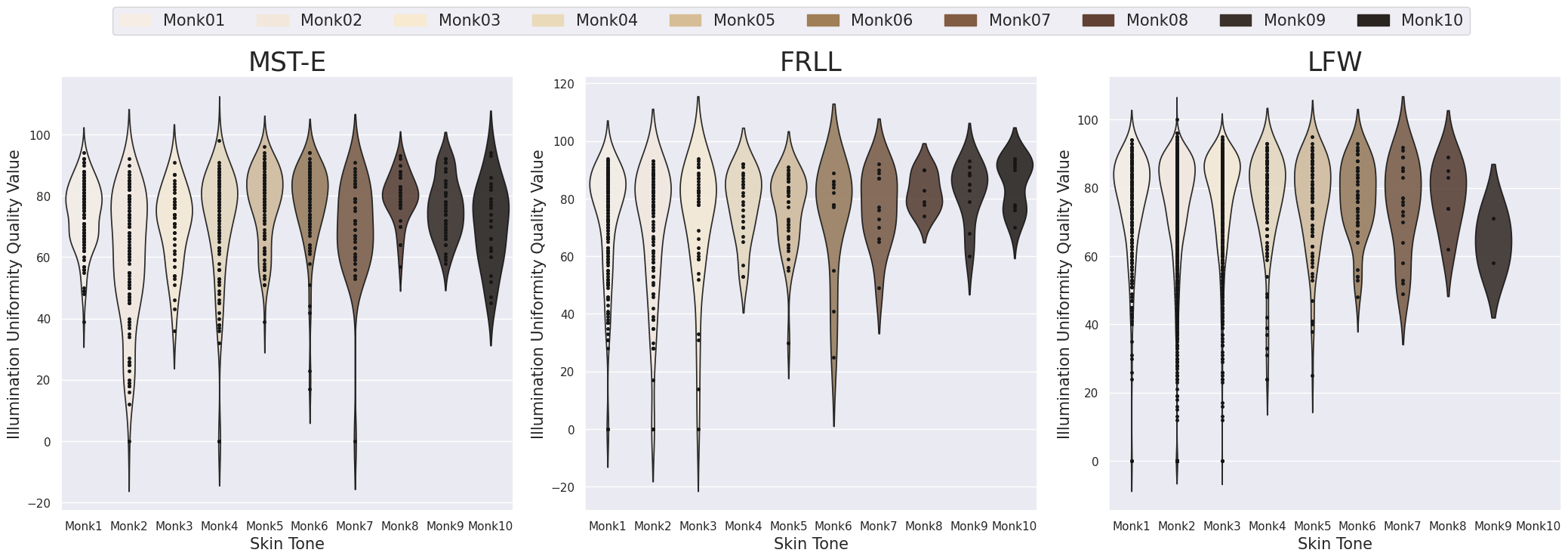}
    \caption{Illumination Uniformity quality value distributions across the MST 10 skin tone scale.}
    \label{fig:skintone-illuni}
\end{figure*}

\begin{figure*}[h]
    \centering
    \includegraphics[width=\linewidth]{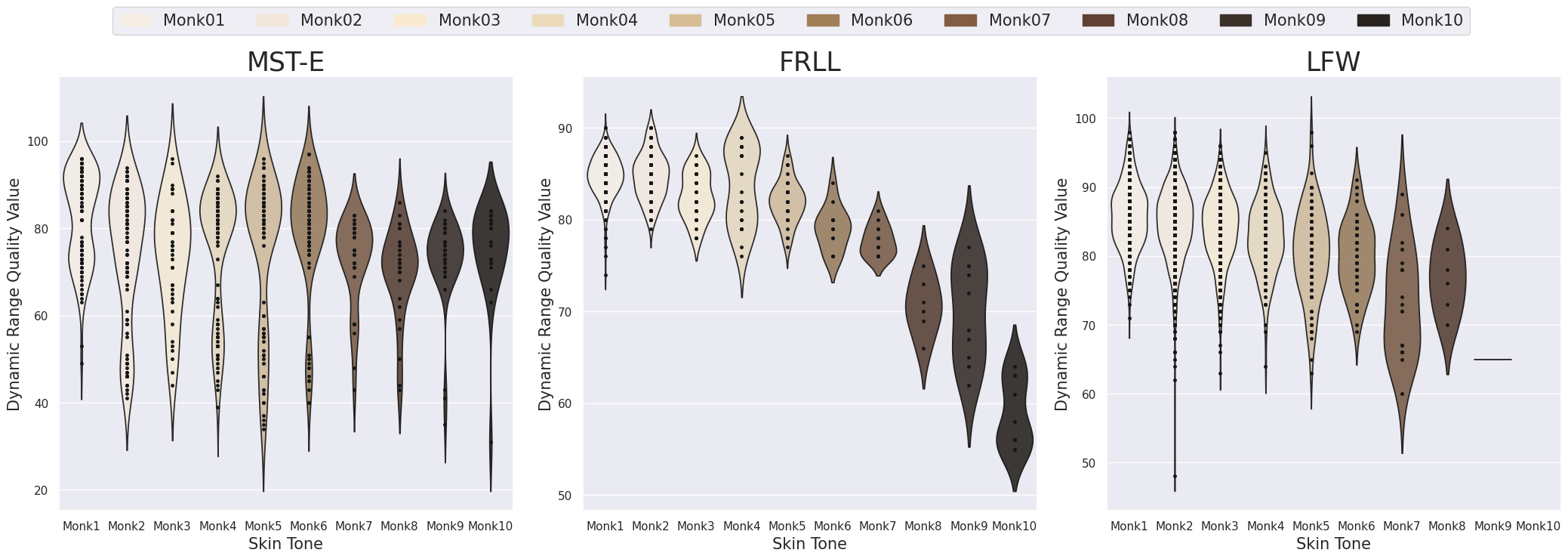}
    \caption{Dynamic Range quality value distributions across the MST 10 skin tone scale.}
    \label{fig:skintone-dynamic}
\end{figure*}

\begin{figure*}[h]
    \centering
    \includegraphics[width=\linewidth]{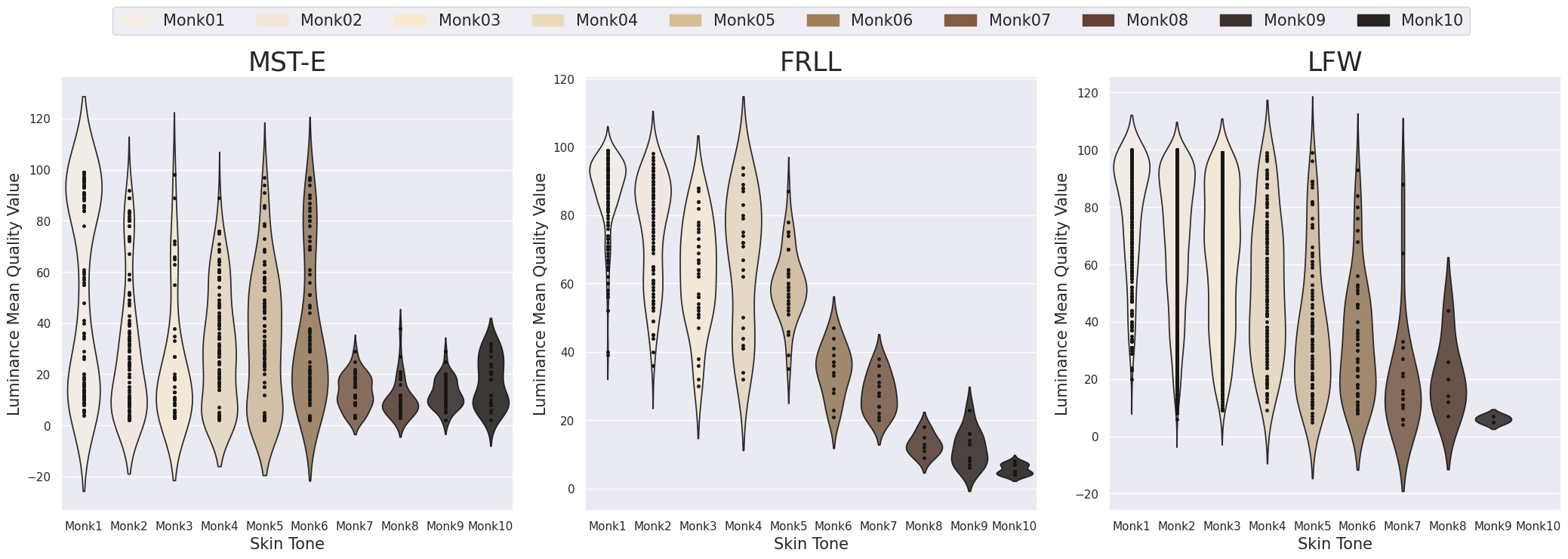}
    \caption{Luminance Mean quality value distributions across the MST 10 skin tone scale.}
    \label{fig:skintone-luminance}
\end{figure*}

We evaluate the FIQA measures on the MST-E, FRLL, and LFW datasets where the ground truth skin tone labels are available. As shown in Figure \ref{fig:skintone-uqa}, the score distributions of the unified quality score do not show any noticeable differences between the various skin tone groups. The scores are rather distributed along the same value ranges on each dataset, with a higher concentration around the median values. The score distributions for most of the other quality measures show rather the same behavior where there are not clear differences between the groups. We show one such example for the illumination uniformity measure in Figure \ref{fig:skintone-illuni}. However, the distributions of two quality measures show clear differences in the results for different skin tone groups. Figure \ref{fig:skintone-dynamic} shows the distributions of the dynamic range quality values. It is clear from the results that lighter skin tones are getting relatively higher quality values. This effect is mildly noticeable in LFW but clearly visible in FRLL. In MST-E, the distributions are almost split into two groups, with the lighter skin tone group having relatively higher quality values. Another quality measure where skin tone is having a very noticeable effect on the quality values is the luminance mean. Figure \ref{fig:skintone-luminance} shows the distributions of the luminance mean quality values. The results of the MST-E dataset are clearly split into the same two groups as for the dynamic range, but with a much larger gap. The two peeks in the distribution are understandable given that MST-E explicitly features images under good and bad illumination settings. The effect of skin tone on this measure is also much more noticeable on FRLL and LFW. On both datasets, the distributions are clearly becoming more concentrated toward lower quality values as the skin tone gets darker.

\subsection{Age}

\begin{figure*}[h]
    \centering
    \includegraphics[width=\linewidth]{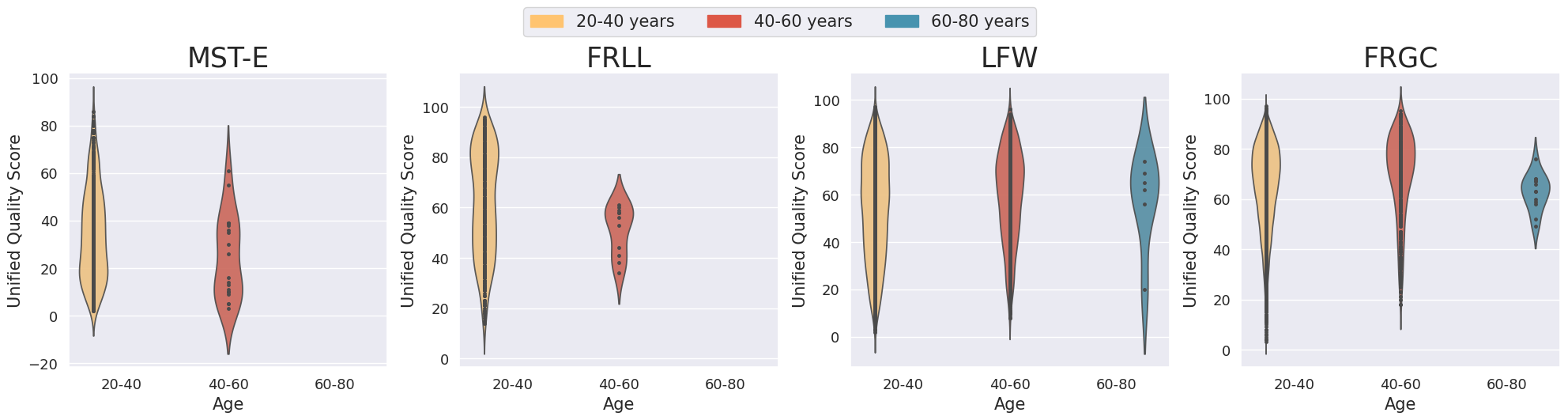}
    \caption{Unified quality score distributions across the 3 age groups. MST-E and FRLL have no subjects in the age group 60-80.}
    \label{fig:age-uqs}
\end{figure*}

We evaluate the FIQA measures on all four datasets. We divide the age label into age groups and retain only the groups that have sufficient representation in one or more datasets. These are age groups: 20–40, 40–60, and 60–80. The evaluation results for all measures show no clear differences between the three age groups. Hence, we choose to show only the distributions of the unified quality scores in Figure \ref{fig:age-uqs}.

\subsection{Gender}

\begin{figure*}[h]
    \centering
    \includegraphics[width=\linewidth]{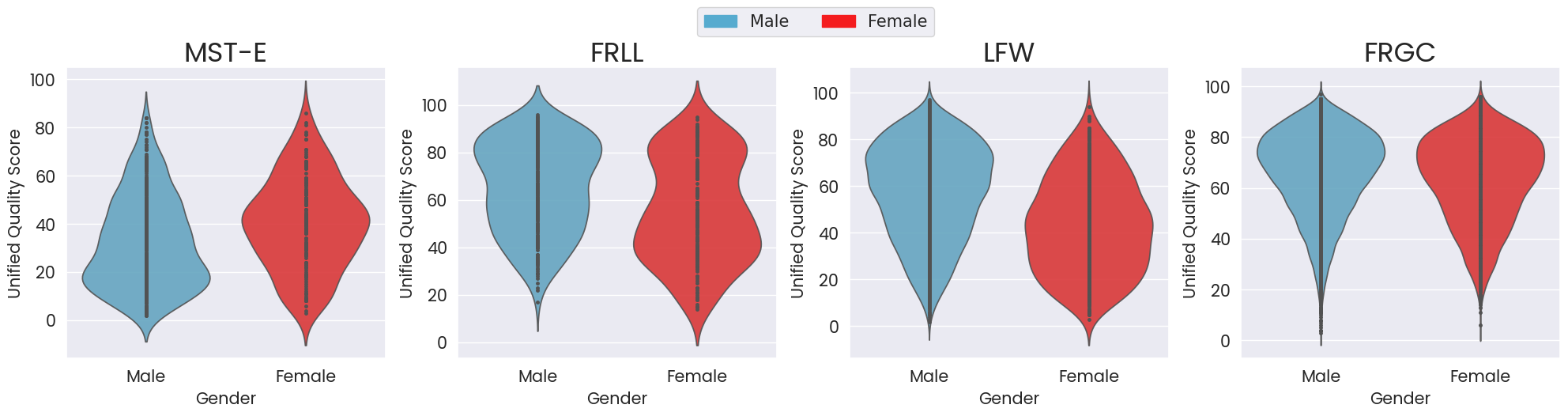}
    \caption{Unified quality score distributions across the 2 gender groups.}
    \label{fig:gender-uqs}
\end{figure*}

\begin{figure*}[h]
    \centering
    \includegraphics[width=\linewidth]{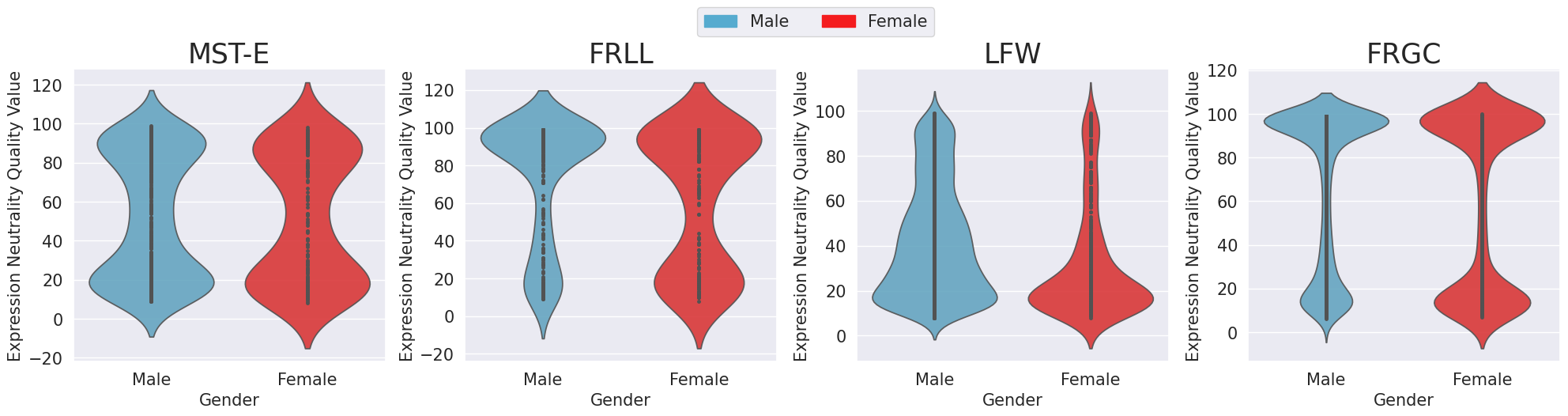}
    \caption{Expression Neutrality quality value distributions across the 2 gender groups.}
    \label{fig:gender-expression}
\end{figure*}

The term \textit{gender} refers to a classification based on social, cultural, or behavioral factors as per the international standard ISO/IEC 2382-37 on biometrics vocabulary \cite{ISO-IEC-2382-37-220330}. In our study, we confine gender to two genders only, given that the ground truth labels and the face attribute classification models report gender in terms of male and female only. Similar to the results for age, there are no clear differences between the two genders in any of the evaluated measures. Figure \ref{fig:gender-uqs} shows the distributions for the unified quality scores, and as evident from the results, the distributions are very similar, with slightly more concentration of higher quality values for the male gender on FRLL and LFW, but on the other hand, slightly more concentration of lower quality values on MST-E. The results on FRGC are rather identical. Figure \ref{fig:gender-expression} shows the quality value distributions for the expression neutrality measure. It is also evident that the distributions are rather identical across the four datasets. The two peeks are also understandable, given that the datasets explicitly feature images of neutral and non-neutral expressions.

\section{Discussion}
\label{sec:discussion}

The findings of the study are rather promising, because unlike what one might expect given the documented demographic bias in facial biometric systems, most FIQA algorithms, studied over four different datasets, have not demonstrated any substantial differences in their results across the four demographic variables. The only two measures that have shown variations in their results on the skin tone variable are the luminance mean and the dynamic range. While it might be expected that these two aspects are different for individuals with different skin tones, it is worth noting that the FIQA algorithms are supposed to produce quality values that reflect how good a given image is with regard to the aspect assessed by the algorithm. Hence, it is not acceptable that these algorithms have variations in their outcomes only due to differences in skin tone.
\section{Conclusion}
\label{sec:conclusion}

In this work, we presented a study on the less studied field of bias in face image quality assessment (FIQA). We conducted a comprehensive evaluation of demographic differentials in all quality measures included in the most recent version of the ISO/IEC 29794-5 international standard. We studied variations in these measures across three demographic variables: age, gender, and skin tone. We studied skin tone differentials on the newly introduced Monk Skin Tone Scale, used its reference labeled dataset, and manually labeled subjects in two other datasets to make sure the results are more reliable. The results of the study give us more confidence when deploying these algorithms as part of online identity management and document issuance applications and point to the specific algorithms that need some attention to alleviate the detected bias.

\section*{Acknowledgment}
\label{sec:acknowledgment}

This work was supported by the European Union's Horizon 2020 Research and Innovation Program under Grant 883356.

\bibliography{main}

\end{document}